\documentclass[10pt,twocolumn,letterpaper]{article}

\usepackage[accsupp]{axessibility}  
\usepackage{wacv}
\usepackage{times}
\usepackage{epsfig}
\usepackage{graphicx}
\usepackage{amsmath}
\usepackage{amssymb}
\usepackage{booktabs}
\usepackage{cuted}
\usepackage{capt-of}

\def\wacvPaperID{363} 

\wacvapplicationstrack 

\wacvfinalcopy 

\ifwacvfinal
\usepackage[breaklinks=true,bookmarks=false]{hyperref}
\else
\usepackage[pagebackref=true,breaklinks=true,colorlinks,bookmarks=false]{hyperref}
\fi

\begin{document}

\title{Interactive Image Manipulation with Complex Text Instructions} 

\author{Ryugo Morita, Zhiqiang Zhang, Man M. Ho, Jinjia Zhou \\ 
Hosei University\\ 
Tokyo, Japan\\ 
{\tt\small ryugo.morita.7f@stu.hosei.ac.jp}
}

\maketitle
\thispagestyle{empty}

\begin{strip}
\begin{center}
    \includegraphics[width=0.75\linewidth]{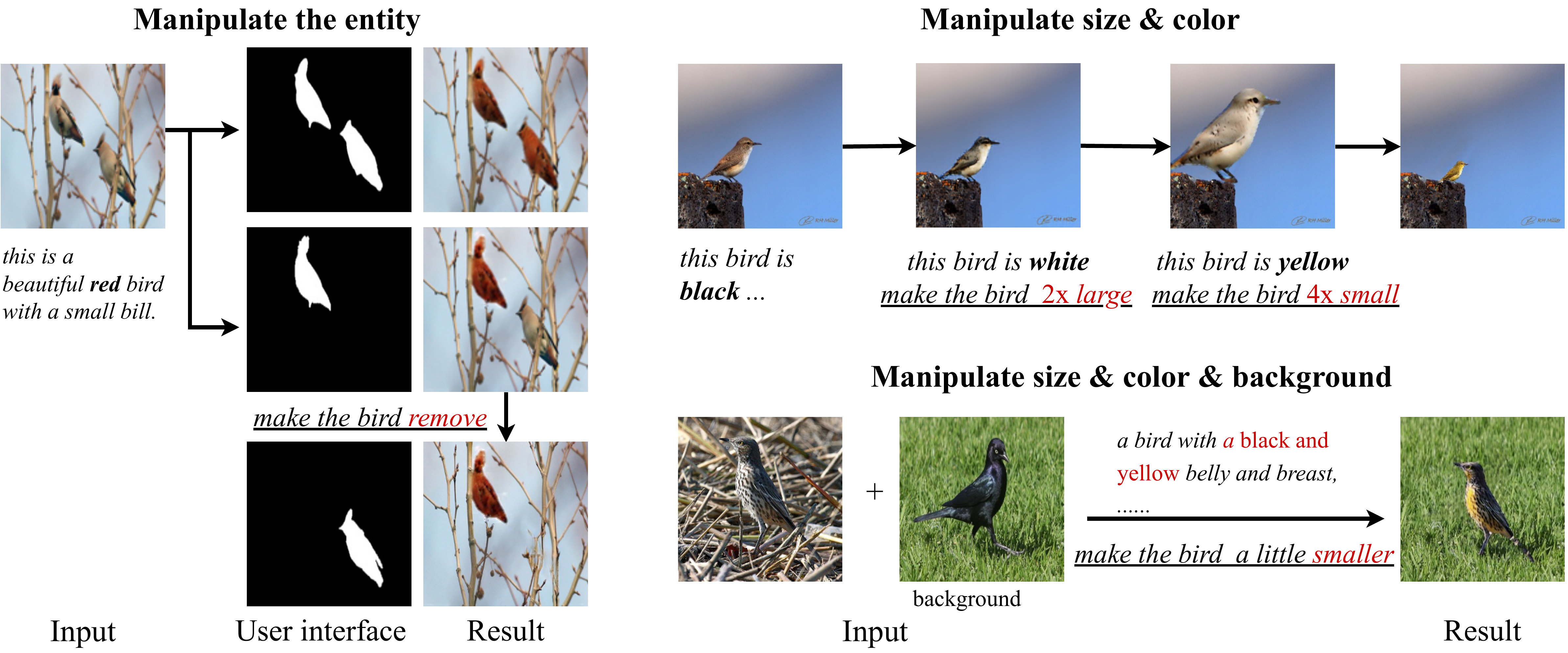}
\end{center}
\captionof{figure}{
       Given an image and text instruction that reveals desired modifications to the image, our method first tries to understand the image and localize where should be modified, then makes appropriate manipulations.
       In addition, our network design also allows users to adjust the affected area and add/redo image manipulations for undesired results.
       As an advantage, this work provides effective image manipulations with high controllability, such as changing an object's attributes (e.g., colors and texture), enlarging, dwindling, removing objects, and replacing the background.
       }
\label{fig:overview}
\end{strip}
%
\begin{abstract}
Recently, text-guided image manipulation has received increasing attention in the research field of multimedia processing and computer vision due to its high flexibility and controllability.
Its goal is to semantically manipulate parts of an input reference image according to the text descriptions.
However, most of the existing works have the following problems:
(1) text-irrelevant content cannot always be maintained but randomly changed,
(2) the performance of image manipulation still needs to be further improved,
(3) only can manipulate descriptive attributes.
To solve these problems, we propose a novel image manipulation method that interactively edits an image using complex text instructions. 
It allows users to not only improve the accuracy of image manipulation but also achieve complex tasks such as
enlarging, dwindling, or removing objects and replacing the background with the input image.
To make these tasks possible, we apply three strategies. 
First, the given image is divided into text-relevant content and text-irrelevant content. Only the text-relevant content is manipulated and the text-irrelevant content can be maintained. 
Second, a super-resolution method is used to enlarge the manipulation region to further improve the operability and to help manipulate the object itself. 
Third, a user interface is introduced for editing the segmentation map interactively to re-modify the generated image according to the user's desires.
Extensive experiments on the Caltech-UCSD Birds-200-2011 (CUB) dataset and Microsoft Common Objects in Context (MS COCO) datasets demonstrate our proposed method can enable interactive, flexible, and accurate image manipulation in real-time. 
Through qualitative and quantitative evaluations, we show that the proposed model outperforms other state-of-the-art methods.
\end{abstract}

\section{Introduction}

Image manipulation aims to modify some aspects of a given image, and the manipulation content includes low-level color or texture to high-level semantics to match the user's preference. It has a wide range of applications in education, image editing, and video game development. In recent years, with the rise of artificial intelligence, automated image manipulation research, including image inpainting\cite{Wang_2021_CVPR}\cite{Li_2020_CVPR}, image colorization\cite{Bahng_2018_ECCV}, style transfer\cite{Kalischek_2021_CVPR}\cite{liu2021adaattn}, and domain or attribute transformation \cite{Chen_2019_CVPR}, has gained widespread attention and made significant progress. Despite achieving stunning results, most studies focus on specific problems and require specialized knowledge, resulting in poor practicality. To address this issue, By using text information that is closely related to people's lives to manipulate the content of the corresponding image region, text-guided image manipulation implements a user-friendly way of image manipulation.

With the rapid development of GAN \cite{goodfellow2014generative} technology, text-guided image manipulation has achieved encouraging results. SISGAN \cite{dong2017semantic} and TAGAN \cite{nam2018text} have achieved decent image manipulation results using a GAN-based encoder-decoder structure and a text-adaptive network, respectively. However, the image results generated by these methods didn't perform well in detail synthesis. In contrast, ManiGAN \cite{li2020manigan} achieves better detail manipulation and makes the final image manipulation result more realistic. Nevertheless, on the one hand, ManiGAN will not only modify the text-relevant content during the manipulation process but also modify the text-irrelevant content, which is not expected. On the other hand, its overall manipulation performance still has room for further improvement. In summary, these existing text-guided image manipulation methods mainly suffer from the following problems:
(1) the content not related to the text description is randomly changed,
(2) the performance of image manipulation still needs to be further improved,
(3) only can manipulate descriptive attributes, lacking specific actions such as removing, enlarging, or dwindling an object.
The root of these problems is that it is difficult to accurately recognize the target content.

\begin{figure*}[t]
\begin{center}
   \includegraphics[width=0.8\linewidth]{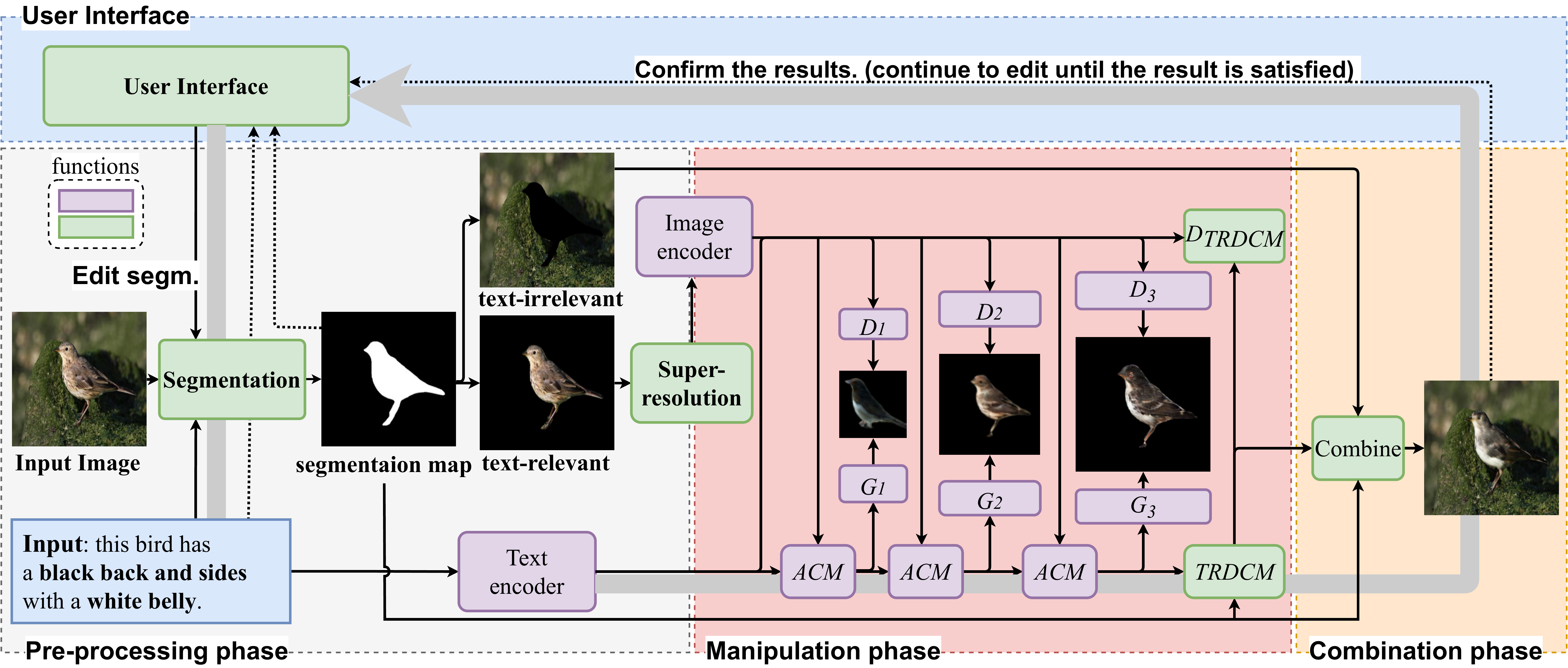}
\end{center}
   \caption{The overview of our proposed model. It allows users to edit the segmentation map automatically.}
\label{fig:model}
\end{figure*}

To solve these problems, we divide the image manipulation task into three phases: pre-processing phase, manipulation phase, and combination phase. In the pre-processing phase, we propose a segmentation network to obtain the segmentation map (as shown in Fig. \ref{fig:model}). The white area in the segmentation map represents the content that needs to be manipulated, and the black area represents the content that does not need to be modified. Based on the segmentation map, the text-relevant content and the text-irrelevant content can be accurately determined. In the manipulation phase, the text-relevant content is manipulated according to the semantic information of the input text, and the text-irrelevant content remains unchanged. In the combination phase, the content of the text-relevant region after manipulation is fused with the text-irrelevant region that keeps the content unchanged to form the final image manipulation result. The whole process is first to determine the text-relevant content and the text-irrelevant content, then only modify the text-relevant content, and then put the modified result back to the original position. 

The above process can solve the first problem mentioned earlier. At the same time, since only the text-relevant content is modified and the text-irrelevant content is retained, the manipulation network can pay more attention to the modification of the text-relevant content, thereby further improving the quality of manipulation. In addition, since the text-irrelevant content is inherently high-quality, the image manipulation results after fusion will have higher quality due to the high quality of text-relevant content and text-irrelevant content.
This solves the aforementioned second problem. However, there are still two issues in the above process. One is that it cannot be guaranteed that the pre-processing phase can obtain an accurate segmentation map. If the segmentation map is inaccurate, the quality of the final manipulation result will be poor. Another is that for the third problem, the above process cannot be solved. For these two issues, we propose a user-interactive segmentation map editing method and use a super-resolution network to solve them. The user-interactive segmentation map editing method allows the user to manually edit the segmentation map to obtain more accurate segmentation map results to ensure the high quality of the final manipulation results. The super-resolution network is used before the image manipulation phase. It can enlarge and dwindle the text-relevant content  so that the ways of our image manipulation are more flexible and diverse. In addition to the above content, to further improve the practicability of our manipulation method and the flexibility of manipulation results, we allow the user to input two original images, select the text-relevant content from one, and retain the text-irrelevant content from the other. However, since the shapes of the text-relevant regions of the two images are mostly different, this leads to holes in the combination stage. Therefore, we adopt the image inpainting methods \cite{yi2020contextual}\cite{ulyanov2018deep} to fill these holes to synthesize high-quality image manipulation results.

Our contributions are as follows:
\begin{itemize}
    \item We propose a novel and effective text-guided image manipulation method. By accurately detecting and manipulating text-relevant content, we achieve better image manipulation results.
    \item In order to improve the detection accuracy of the text-relevant region and the user-friendly of the whole method, we propose a user-interactive segmentation map editing method, which allows users to manually edit the detected segmentation map to achieve a more accurate detection effect to make the final image manipulation result more conform to the user's expectations.
    \item To make the proposed method more flexible and diverse, we introduce a super-resolution network and an image inpainting network to achieve richer manipulation operation forms such as enlarge, dwindle, background replacement, and so on.
    \item Experiments on the CUB \cite{wah2011caltech} and MS COCO \cite{DBLP:journals/corr/LinMBHPRDZ14} datasets demonstrate the effectiveness and superiority of our proposed method.
\end{itemize}

\section{Related Works}
\textbf{Text-to-Image Synthesis.} The success of text-to-image synthesis by generative adversarial networks (GAN) \cite{goodfellow2014generative} attracted a lot of attention.
Reed et al. \cite{reed2016generative} pioneered the end-to-end text-to-image research base on GAN. Following that, the Generative Adversarial What-Where Network (GAWWN) \cite{reed2016learning} was shown to synthesize content at specified locations, resulting in superior quality images. Moreover, StackGAN \cite{zhang2017stackgan} and StackGAN++ \cite{zhang2018stackgan++} architecture aim to generate realistic images using multi-stage structures with generators and discriminators at each stage. Based on this, various improvements have been made with the aim of improving realism \cite{xu2018attngan} \cite{zhu2019dm} \cite{zhang2021draw}, semantics \cite{qiao2019mirrorgan} and diversity \cite{bodla2018semi}\cite{dash2017tac}.

\textbf{Text-guided Image Manipulation} aims to use text to modify text-relevant content in the input image and maintain text-irrelevant content.
Dong et al.\cite{dong2017semantic} proposed an encoder-decoder architecture to edit an image according to a given text. Then, TAGAN \cite{nam2018text} aimed to preserve the text-relevant content by introducing a text-adaptive discriminator, which can give the generator finer training feedback, to achieve better manipulation results. More recently proposed ManiGAN \cite{li2020manigan} introduced the multi-stage structures with generators and discriminators at each stage for high-quality manipulation results. In addition, StyleCLIP \cite{StyleCLIP} achieved the style transfer effect of images based on text-driven. However, this work is more applied to style transfer of facial content, which leads to its general practicability. Besides, the common problem with the aforementioned methods is that they do not work well in preserving text-irrelevant content. To solve this problem, Tomoki et al. \cite{haruyama2021segmentation} proposed the approach to introducing a segmentation function to reduce the loss of text-irrelevant content.
This work is close to our proposed method. However, this method can only reduce the modification degree of the background as much as possible and cannot avoid the modification of the background content.
Different from \cite{haruyama2021segmentation}, we design an enhanced architecture that divided an image into text-relevant content and text-irrelevant content and achieved zero loss for text-irrelevant content. Furthermore, we achieve challenging tasks such as enlarging, dwindling, and removing an object, which other existing methods can not do.
Additionally, the proposed method introduces an interface that can edit the segmentation interactively so that our method has excellent flexibility and practicality.
\section{Methods}
\label{sec:pagestyle}
As shown in Fig. \ref{fig:model}, the inputs of the proposed system are an input image $I$ and a text instruction $S$. 
The target is to edit the content associated with $S$, preserve the content not associated with $S$, and generate a corresponding image $I^{\prime}$.
Firstly, the segmentation network is applied to divide the input image into two regions: text-relevant content $I_{tr}$ and text-irrelevant content $I_{ti}$.
Secondly, we input text-relevant content in the super-resolution network \cite{zhang2021designing} to obtain the refined text-relevant content $I_{tr\_SR}$.
Then, we adopt the ManiGAN \cite{li2020manigan} as the basic framework to edit the refined text-relevant content according to the text instruction.
Finally, the output image $I'$ is obtained by combining the edited text-relevant and the original text-irrelevant content into the combination network.
Moreover, in order to more accurately detect the content of text-relevant regions, we propose a user interface that allows users to edit the segmentation map detected before so that the final manipulation result is more conform to the user's expectation.
Fig. \ref{fig:model} shows our entire system divided into three phases: pre-processing phase, manipulation phase, and combination phase. Next, we introduce the specific content of each phase one by one.
\subsection{Pre-processing phase}
\label{ssec:segmentaiton}
\textbf{Segmentation network.} Existing works are not able to recognize the target object in the text description, causing current text-guided image manipulation methods to suffer problems such as randomly changing the content of the text-irrelevant region.
To solve this issue, we introduce a segmentation network. 
This network allows linking the text with the corresponding region information in the image.
It takes the input image $I$ and the text instruction $S$ as the input, then links text-region information by performing two tasks simultaneously.
The first task is to obtain the segmentation mask $M_{seg}$. We detect all objects with re-trained Deeplabv3 \cite{chen2018encoder} from scratch on the CUB and COCO datasets, respectively, to achieve this task.
The second task is to acquire the candidate class label $W$ of objects, we employ pre-trained YOLOv4 \cite{DBLP:journals/corr/abs-2004-10934} to achieve this task.
When finishing the above two tasks, the cosine similarity between the words in the obtained candidate class label $W$ and the noun words in the text instructions $S$ is calculated to obtain the class information $c$ of the most relevant object.
\begin{equation}
  c = \text{max}(\text{Similarity}(W, S))
\end{equation}
The region corresponds to text-relevant class $c$ is selected from $W$ and then multiplied with $M_{seg}$ to obtain the text-relevant mask $M_{tr}$.
\begin{equation}
M_{tr} = M_{seg} * W(c)
\end{equation}
We mask the input image $I$ by using text-relevant mask $M_{tr}$ to obtain the text-relevant and text-irrelevant content:
\begin{equation}
I_{tr}, I_{ti} = \text{Mask}(M_{tr}, I)
\end{equation}

\textbf{Super-Resolution network.} When the text-relevant content is in a small region or when modifying the detailed information (e.g., eyes), the existing works are unable to achieve sufficient accurate positioning detection resulting in a poor manipulation effect. Besides, the existing methods can only modify the manipulation at the pixel level and cannot achieve manipulation effects such as object enlarging and dwindling.
To solve the above issues, we adopt the super-resolution network.
First, we crop the object region that needs to be manipulated in the text-relevant content.
Then, it is up-sampled by an image super-resolution network (pre-trained BSRGAN \cite{zhang2021designing}) to obtain the super-resolution text-relevant content $I_{tr\_SR}$. 
The super-resolution network can zoom in on the small region and detailed information to provide more effective information for the subsequent manipulation phase and achieve better manipulation results. Besides, it allows enlarging and dwindling of the object so as to achieve more manipulation effects.
Furthermore, this network is first used to make the object in the largest size that would fit in the image. And then, in the manipulation phase, the size of the object is modified by receiving text instructions and matched to that size to generate the object content.
\begin{figure}[t]
\begin{center}
   \includegraphics[width=\linewidth]{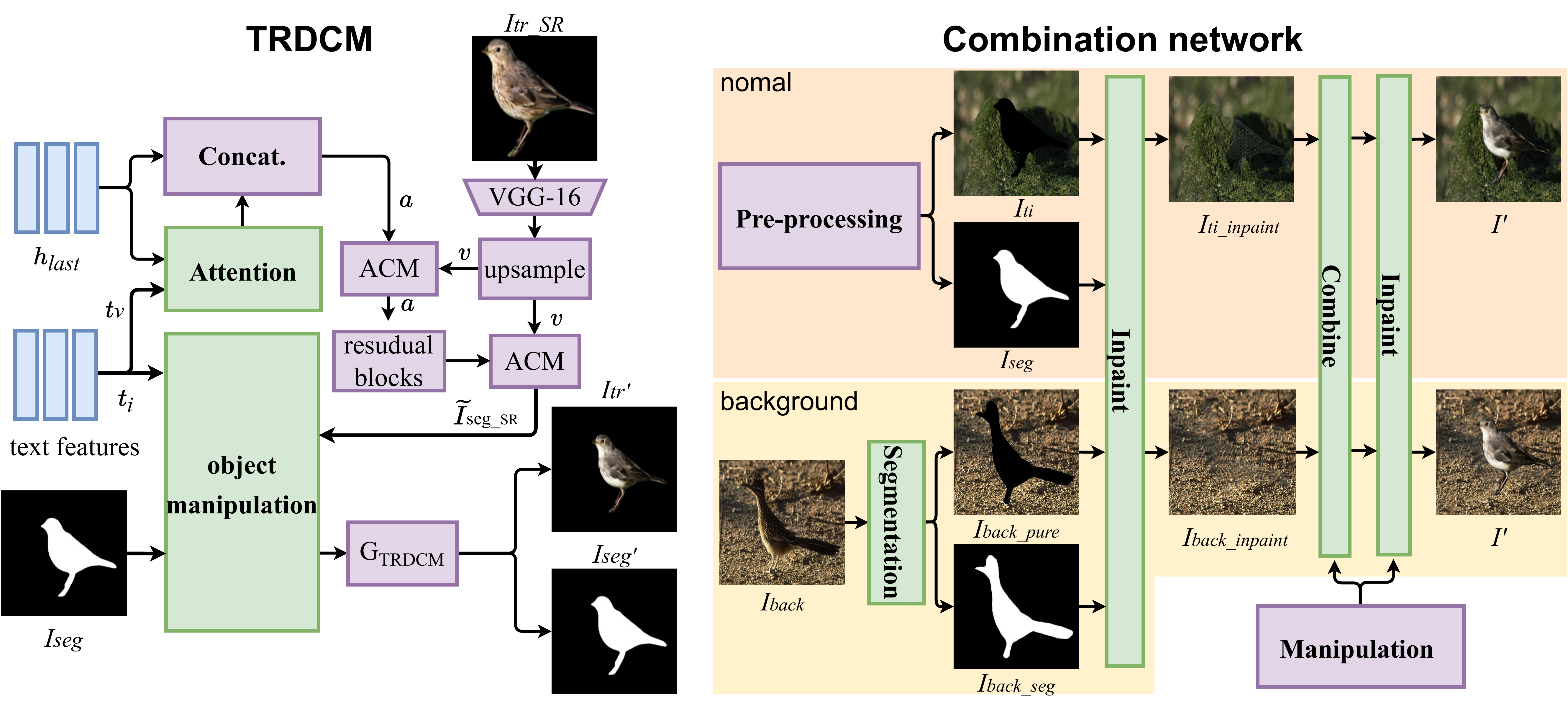}
\end{center}
   \caption{The detailed content of TRDCM and Combination network is shown above.}
\label{fig:detail}
\end{figure}
\begin{figure*}[tb]
\begin{center}
   \includegraphics[width=0.8\linewidth]{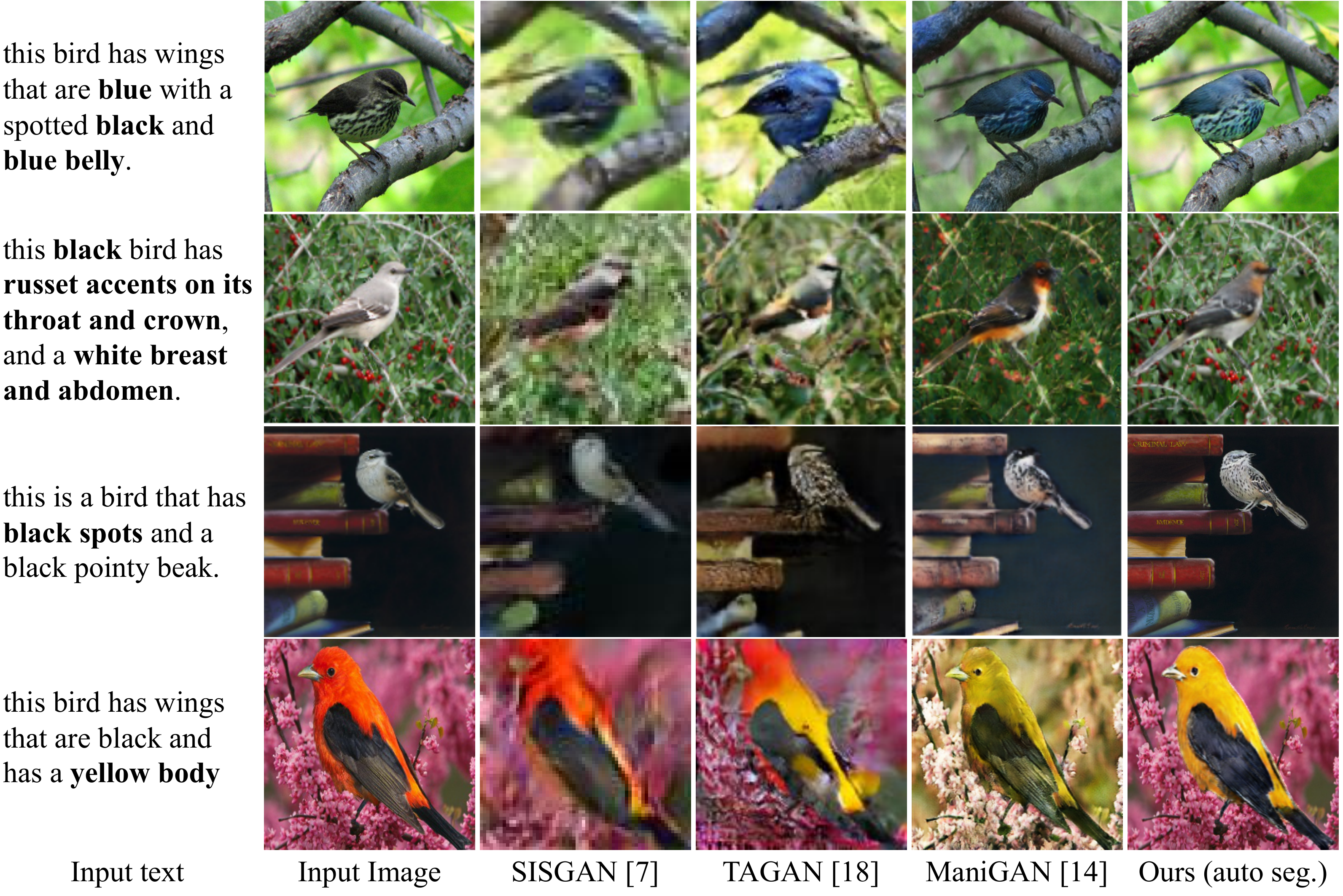}
\end{center}
   \caption{The qualitative comparison of the proposed ours(auto seg.) and related works on the CUB dataset. Existing methods lack details (e.g., eyes, patterns, etc.), and fail to preserve the text-irrelevant content. In contrast, by adopting semantic map information and the super-resolution method, this method successfully keeps text-irrelevant content and generates better text-relevant content.}
\label{fig:qualitative_wo_interface}
\end{figure*}
\subsection{Manipulation phase}
\label{ssec:manipulation}
In the manipulation phase, we refer to the affine combination module (ACM) and the detail correction module (DCM) in ManiGAN\cite{li2020manigan}. Unlike ManiGAN, we propose a novel Text-Relevant DCM (TRDCM) to replace DCM. The contents of ACM and TRDCM in our method are as follows.

\textbf{ACM.} Different from ACM in ManiGAN, ACM in our method processes the features by the super-resolution network. Since the features of the super-resolution network belong to the text-relevant region, our manipulation results can more conform to the semantic information of the text and do not change the content of the text-irrelevant region content. In contrast, ManiGAN's ACM processes the features of the entire image, which leads to the problem that the content of the text-irrelevant region also changes.

\textbf{TRDCM.} We do not introduce the user interface during the training processing, so our TRDCM content is consistent with ManiGAN's DCM. However, our TRDCM is quite different from ManiGAN due to the introduction of user interaction in the testing. Its basic structure is shown on the left side of Fig. \ref{fig:detail}. Specifically, it includes four inputs:
(1) the last hidden feature $h_{last}$ from the ACM,
(2) the text features $t$,
(3) the super-resolution text-relevant content $I_{tr\_SR}$ and,
(4) the segmentation map $I_{seg}$ from the pre-processing network.
The text features consist of the word-visual vector $t_{v}$ and the word-instruction vector $t_{i}$ encoded by a pre-trained RNN \cite{xu2018attngan}.
First, the word-visual vector $t_{v}$ is input to the spatial and channel-wise attention \cite{NEURIPS2019_1d72310e} to create the attention features, which are then combined with $h_{last}$ to obtain the intermediate feature $a$. 
Secondly, to obtain the detail visual vector $v$ as the same size as $a$, we upsample the image feature encoded $I_{tr\_SR}$ by the pre-trained VGG-16 \cite{simonyan2014very}. Then, by using the ACM, $v$ and $a$ fused to produce the feature $\tilde{a}$. we refine $\tilde{a}$ with residual block according to \cite{li2020manigan} to obtain the image feature which edit the descriptive information $\tilde{I_{tr\_{SR}}}$. Finally, we concatenate the image feature $\tilde{I_{tr\_{SR}}}$ with word-instruction vector $t_{i}$ to produce the final modified text-relevant image $I^{\prime}_{tr}$. Since the text-relevant content obtained after the super-resolution network processing is the largest size, during the generation process in TRDCM, the segmentation map is used to guide the generation of text-relevant region manipulation result that is consistent with the size of the segmentation map. Furthermore, in order to introduce more manipulation operations (such as object enlarging, dwindling, and removal), we detect keywords in the text in the TRDCM of the testing phase. For example, if the word ``2x large'' or ``4x small'' appears, we will scale the segmentation map accordingly to obtain $I_{seg}^{'}$ and then use it to guide the generated size of the text-relevant region content. If the word ``remove'' appears, we will remove the content of the corresponding region according to the current input segmentation map.

\begin{figure*}[tb]
\begin{center}
    \includegraphics[width=0.8\linewidth]{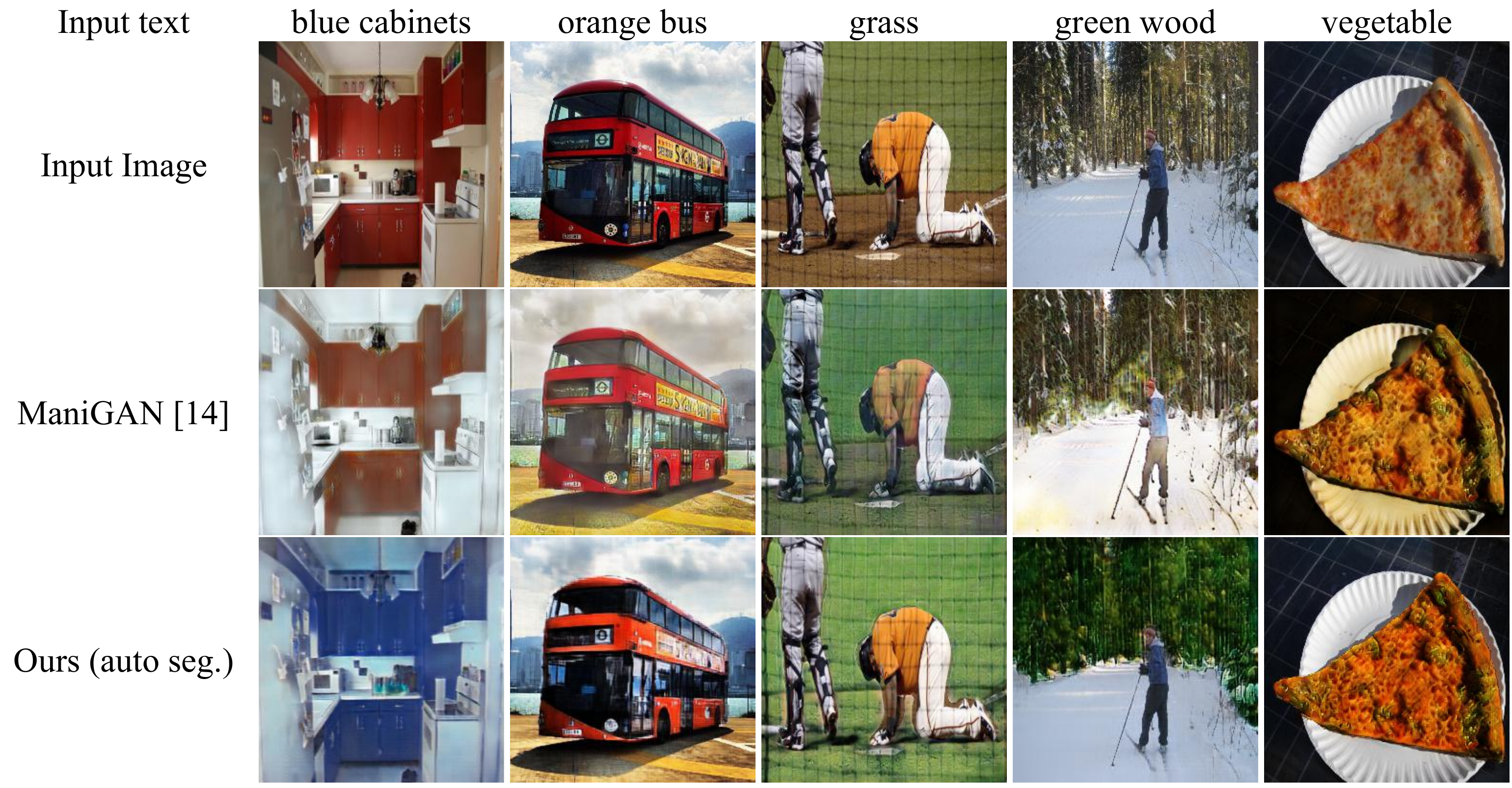}
\end{center}
    \caption{The qualitative comparison of the proposed ours (auto seg.) and related works on the COCO dataset. We can see that ManiGAN manipulates all regions of the image. In contrast, our model is capable of editing images only for a specific object so that it can generate better results.}
\label{fig:qualitative_wo_interface_COCO}
\end{figure*}
\subsection{Combination phase}
\label{ssec:com}
\textbf{Combine network.} We introduce this network (Fig. \ref{fig:detail}) to combine the edited text-relevant content and the original text-irrelevant content to synthesize a plausible image.
It takes three required inputs and one optional input.
(1) the edited text-relevant content $I_{tr}^{'}$ from the manipulation phase,
(2) the original text-irrelevant content $I_{ti}$,
(3) the segmentation map $I_{seg}$ from the pre-processing phase, 
(option) the reference background image $I_{back}$.
First, according to \cite{yi2020contextual}, $I_{ti}$ is inpainted based on the segmentation map to remove the text-relevant content to obtain the inpainted text-irrelevant content $I_{ti\_inpaint}$.
In the background change task, the reference background image $I_{back}$ is input to Deeplabv3 \cite{chen2018encoder} to obtain the pure background image $I_{back\_pure}$ and segmentation map $I_{back\_seg}$. Then, based on the segmentation map, the inpainted background image $I_{back\_inpaint}$ can be obtained.
And then, to obtain the output image $I'$, based on the segmentation map $I_{seg}^{'}$ obtained by the manipulation network, we combine the modified text-relevant content $I_{tr}^{'}$ and the inpainted text-irrelevant content $I_{ti\_inpaint}$, or the inpainted background image $I_{back\_inpaint}$ for the background change task.
Finally, we introduce a function to absorb the color differences in these contents that are caused by processing them separately.
Specifically, the outline of the segmentation map $I_{seg}^{'}$ is extracted to make the mask. By masking this outline for the combined image, we produce the image without the contour region. 
By inpainting the image without the contour region based on \cite{ulyanov2018deep}, the color difference can be eliminated.
%
%
\subsection{Interactive user interface}
\label{ssec:interactive_editing}
As shown in Fig. \ref{fig:mis_seg1}, for some cases, the output segmentation map automatically generated by the network is not perfect. To solve this issue, we adopt the user interface to edit the segmentation map in real-time.
With the user interface, when the initial image manipulation does not work well, accurate image manipulation can be performed by revising the segmentation result.
As shown in Fig. \ref{fig:n_times_modify}, the editing can be done recursively, and it can be corrected and redone many times.

\subsection{Loss function and Training}
\label{ssec:function}
The above parts that need to be trained only include the manipulation phase, and the baseline used in the training process is ManiGAN \cite{li2020manigan}. The pre-processing phase, combination phase, and user interface proposed by our are to solve the problem that the existing methods cannot accurately manipulate the required objects and simultaneously achieve more manipulation operations so that the entire model has better applicability.

In the training process, the overall structure is based on GAN. Following ManiGAN \cite{li2020manigan}, it includes a main module and a TRDCM, where the main module includes three ACM, corresponding to three generators and three discriminators. The TRDCM module includes a generator and a discriminator. The main module and TRDCM are trained individually.

\textbf{Generator objective.} Based on \cite{li2020manigan}\cite{NEURIPS2019_1d72310e}, the generator loss $L_{G}$ consists of an adversarial loss $L_{adv}$, a perceptual loss $L_{per}$, a text-image matching loss $L_{DAMSM}$, and a regularization loss $L_{reg}$.  We show the definition:
\begin{equation}
  L_{reg} = 1-\frac{1}{CHW}||I^{'}_{tr}-I_{tr}|| 
\end{equation}
\begin{equation}
\fontsize{8pt}{0cm}\selectfont
\begin{split}
    L_{G} = L_{adv} + L_{per} + {1 - L_{cor}(I_{tr}^{'}, S)} \\ + L_{DAMSM}(I', S) + L_{reg}
\end{split}
\end{equation}
where $C$, $H$, and $W$ are the number of color channels, the height and width of the input image $I_{tr}$, respectively.

\textbf{Discriminator objective.} Based on \cite{li2020manigan}, the discriminator loss $L_{D}$ consists of an adversarial loss $L_{adv}$ and the text-image correlation loss $L_{cor}$. The discriminator loss $L_{D}$ is defined as follows:
\begin{equation}
\fontsize{8pt}{0cm}\selectfont
\begin{split}
L_{D} = L_{adv} + {1 - L_{cor}(I_{tr}^{'},S)} + L_{cor}(I_{tr}^{'},S^{'})
\end{split}
\end{equation}
where $S^{'}$ is a randomly chosen mismatched textual description from the dataset.

\section{Experiments}
\label{sec:experiments}

We evaluate our method on the CUB and COCO datasets. The CUB dataset \cite{wah2011caltech} contains 11,788 bird images of 200 categories. 8,855 and 2,933 images are used for training and testing, respectively. The COCO dataset \cite{DBLP:journals/corr/LinMBHPRDZ14} contains 123,287 wide range of genre images. 82,783 and 40,504 images are used for training and testing, respectively. Using a user interface does not ensure reproducibility and results in unbiased evaluations. For fair evaluation, we compare our model (auto seg.) with state-of-the-art models in qualitative and quantitative performance.
Our model (auto seg.) means a model that performs image manipulation using a segmentation map automatically detected by the segmentation network, without the user interface to allow segmentation editing.
Based on the Adam optimizer\cite{kingma2014adam}, we train the main module 600 epochs on the CUB dataset and 120 epochs on the COCO dataset, and train the TRDCM module 100 epochs for both datasets.
\subsection{Comparison with state-of-the-art approaches}
\label{ssec:exp_comparison}
\textbf{Quantitative comparison.} In the quantitative experiment, we evaluate the IS \cite{salimans2016improved}, NIMA \cite{talebi2018nima}, and FID \cite{heusel2017gans} on randomly selected images from the CUB and COCO datasets with a randomly chosen text description. Following the ManiGAN, 30,000 images are generated for quantitative evaluation. The results are shown in Table \ref{table:qualitative}. On the CUB and COCO datasets, our method (auto seg.) outperforms on almost all metrics to compare state-of-the-art models, except for the NIMA on the COCO dataset. In the IS, this means that our segmentation network produces highly discriminative images by considering the text-relevant and text-irrelevant content. Furthermore, the NIMA of our method is also high, indicating that our method achieves generating high-quality manipulated images. Furthermore, the excellent FID results indicate that our results have the best fidelity.
\begin{table}[tb]
  \caption{Quantitative comparison results between our method and other existing methods are shown below. We use IS, NIMA, and FID for quantitative comparisons.}
  \label{table:qualitative}
  \centering
  \setlength{\tabcolsep}{3mm}
  \begin{tabular}{l|ccc}
    \hline
     & IS $\uparrow$ & NIMA $\uparrow$ & FID $\downarrow$ \\
    \hline
    \hline
    \multicolumn{4}{c}{CUB}\\
    \hline
    \hline
     SISGAN \cite{dong2017semantic}  & 2.33 & 4.51 & 258.56 \\
     TAGAN \cite{nam2018text}   & 3.05 & 4.70 & 184.79 \\
    ManiGAN \cite{li2020manigan}  & 4.60 & 5.17 & 11.97 \\
    SEGMani \cite{haruyama2021segmentation}  & 4.55 & 4.26 &  - \\
    Ours w/o SR  &  6.30  &  5.25 & 9.56 \\
    Ours (auto seg.)  &  {\bf6.79}  &  {\bf5.32} & {\bf7.15} \\
    \hline
    \hline
    \multicolumn{4}{c}{COCO}\\
    \hline
    \hline
    ManiGAN \cite{li2020manigan}  & 18.20 & {\bf5.19} & 39.74 \\
    Ours w/o SR  &  21.47  &  5.15 & 28.48 \\
    Ours (auto seg.)  &  {\bf21.67}  &  5.16 & {\bf25.60} \\
    \hline
  \end{tabular}
\end{table}

\textbf{Qualitative comparison.} Fig. \ref{fig:qualitative_wo_interface} and Fig. \ref{fig:qualitative_wo_interface_COCO} show the comparison between the previous models and ours (auto seg.).
The comparison results indicate that our model is able to manipulate the image to match the text description.
Moreover, the edited results produced by previous models sometimes are less satisfactory, such as the text-relevant content is not edited and the text-irrelevant content is changed. 
In contrast, because the text-relevant and text-irrelevant region is separated by segmentation, only the text-relevant content is edited in our results. 
A comparison of the details of each image shows that our method is able to preserve the details. Specifically, for the background image, our method is able to retain detailed information such as the name of the book and the surface information of the tree. 
Moreover, previous models often fail to maintain complex information and fail to manipulate the target objects. This problem can be solved by our model because our model is capable of recognizing the target objects automatically, and then only the recognized region is manipulated.
Furthermore, previous models only can accurately manipulate large objects. In contrast, ours can accurately manipulate both large objects and small objects due to the use of a super-resolution network.
\begin{figure}[tb]
\begin{center}
    \includegraphics[width=8cm]{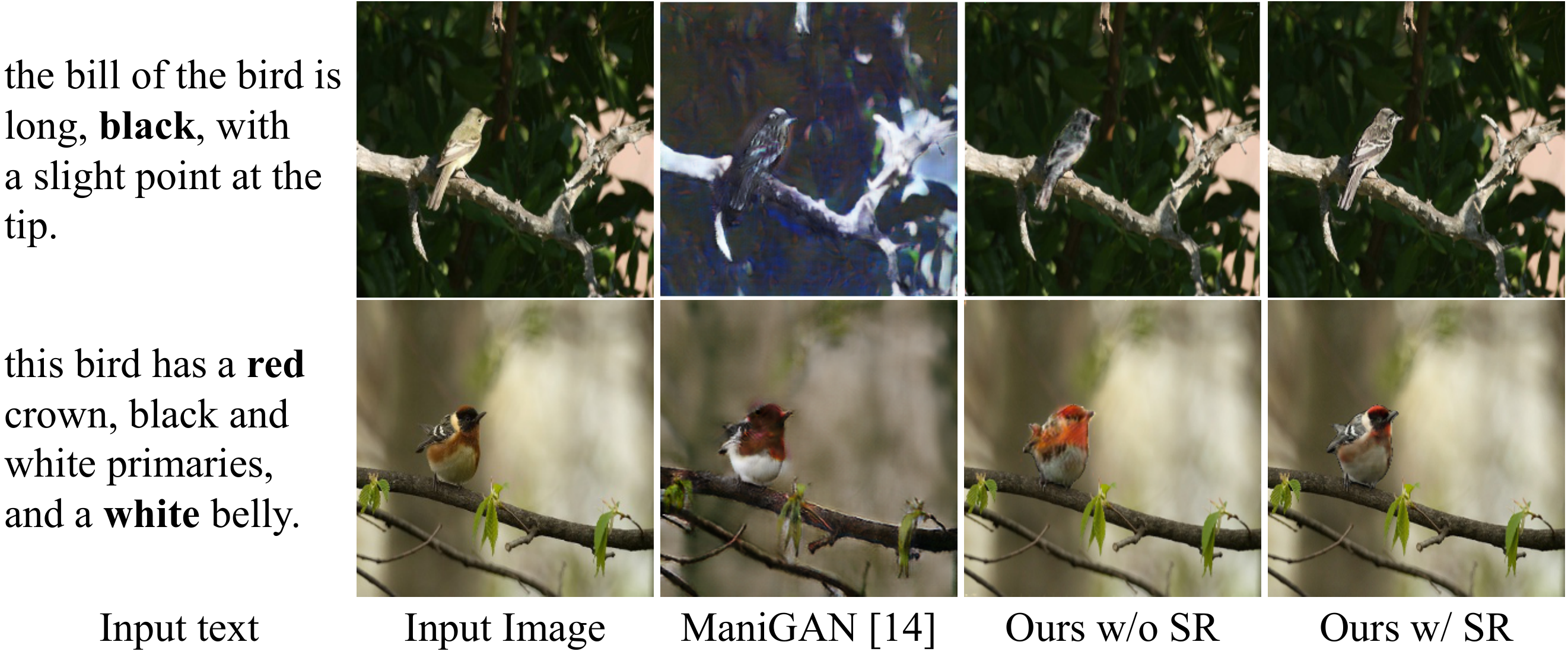}
\end{center}
    \caption{The qualitative comparison of ours w/o super-resolution, w super-resolution, and ManiGAN.}
\label{fig:ablation}
\end{figure}
\begin{figure}[tb]
\begin{center}
    \includegraphics[width=8cm]{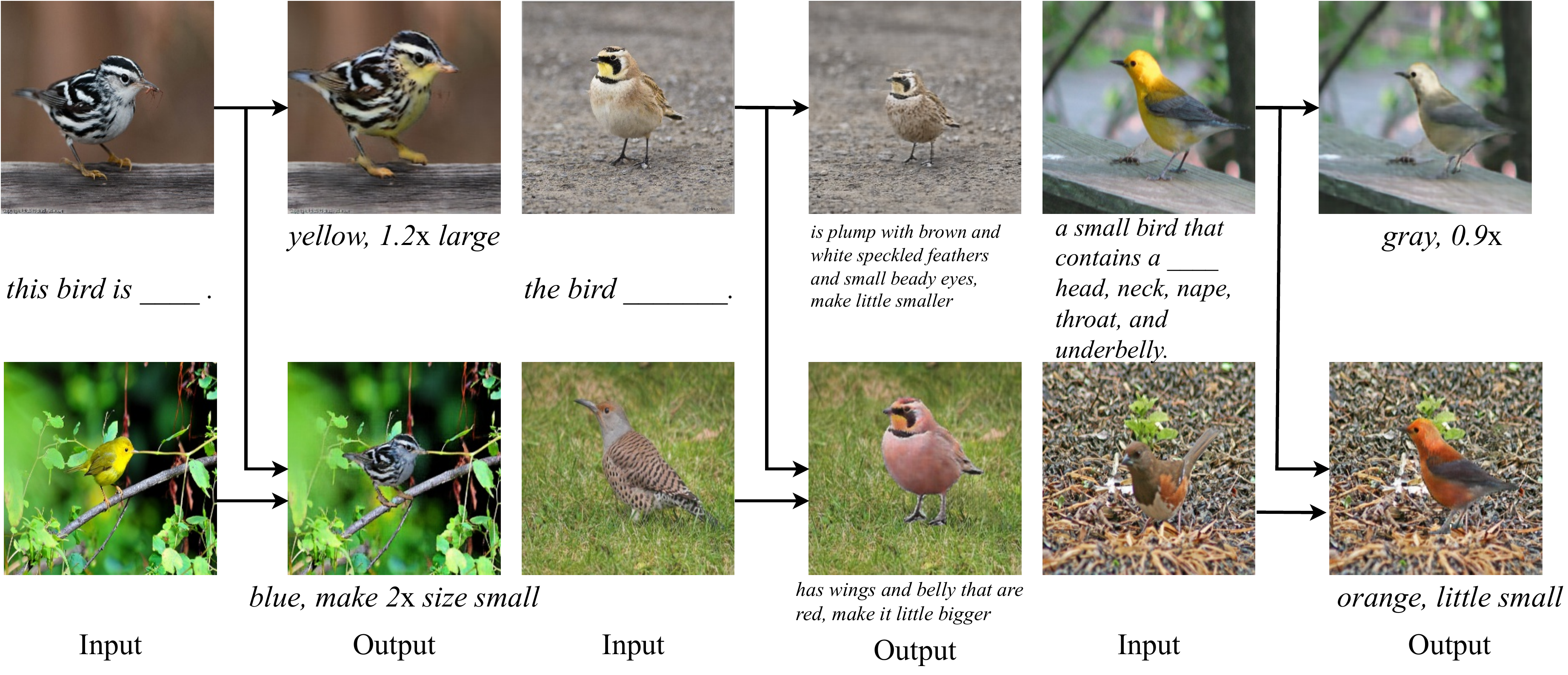}
\end{center}
    \caption{The result of changing the size and the background.}
\label{fig:change_size}
\end{figure}

\subsection{Component Analysis}
\label{ssec:exp_SR}

\textbf{Improvement from super-resolution.} To better understand what has the benefit of our super-resolution network, we visualize the generated images without the super-resolution in Fig. \ref{fig:ablation}.
In the model without super-resolution, detailed information about the bird is missing, while in the model with super-resolution are made detailed information is retained, such as textures and colors. It implies that super-resolution helps retain detailed information in the image.
As shown in Table \ref{table:qualitative}, compared with w/o super-resolution, our method with super-resolution performs better, which proves the effectiveness of adding the super-resolution network.

\begin{figure}[tb]
\begin{center}
    \includegraphics[width=7cm]{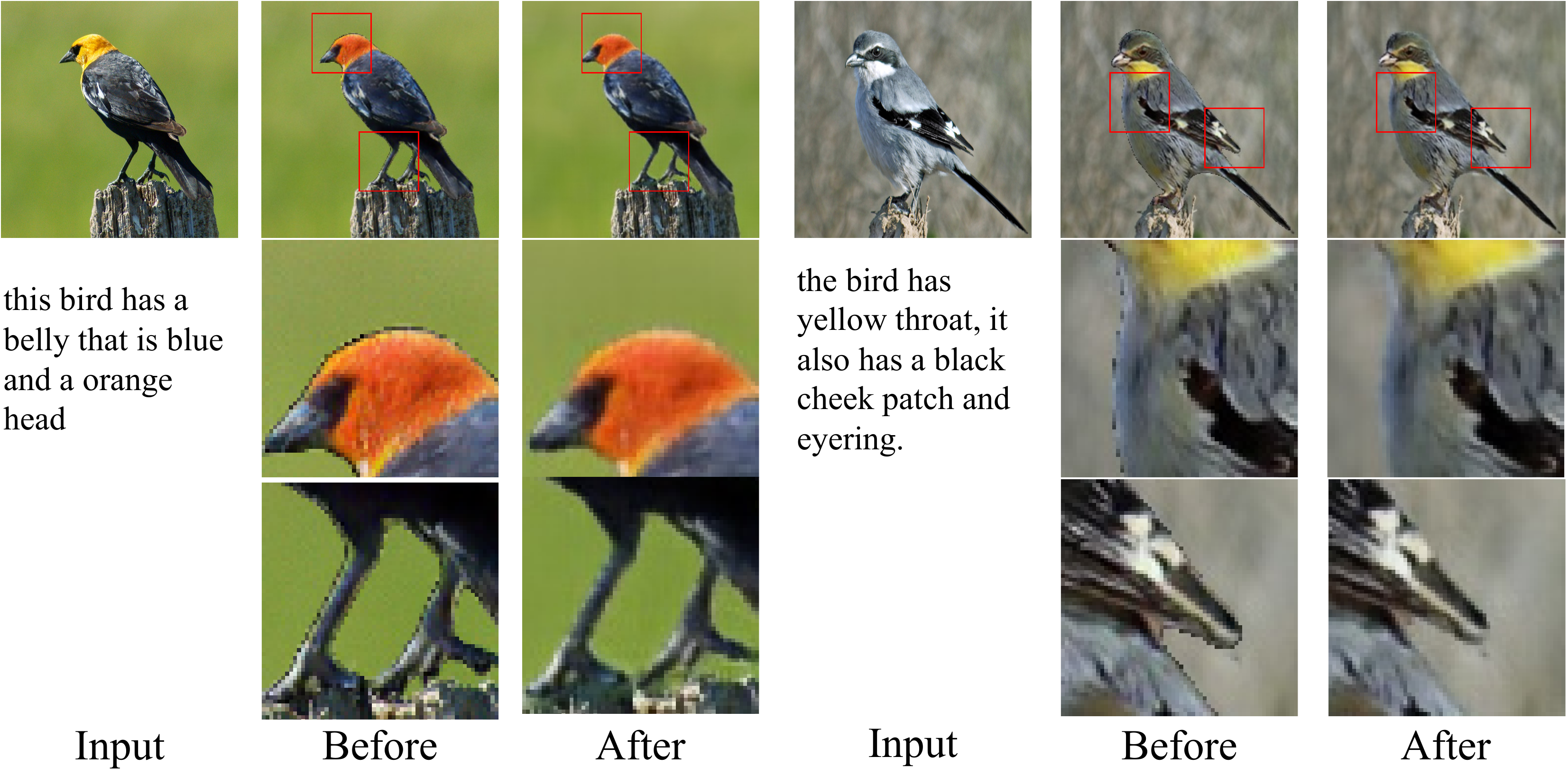}
\end{center}
    \caption{Comparison of the results before and after inpainting.}
\label{fig:inpaint}
\end{figure}

Furthermore, this architecture can also be used to resize the objects in the input image. As shown in Fig. \ref{fig:change_size}, the size of the object can now be changed to any size.
Whether making a small object larger or making a large object smaller, both processes can be achieved in our model.
It's noted that the above operations are not possible to achieve in existing models due to the limitation of only being able to modify the entire image at once.

\textbf{Improvement from Inpaint network.} During the combination phase, holes appear when objects dwindle and background replacement operations are performed. To fill in the hole information and make the final manipulation result more realistic, we employ two image inpainting networks \cite{yi2020contextual}\cite{ulyanov2018deep}. The networks are specifically used for content repair at two levels. The first one \cite{yi2020contextual} is to repair the hole information left in the original image after segmenting the text-relevant content. As shown in Fig. \ref{fig:change_size}, due to effective content repair, our method is able to synthesize satisfactory results when encountering the operations of object enlarging, dwindling, and background replacement. The second \cite{ulyanov2018deep} is to fix the color difference phenomenon. We find that when the object is divided, manipulated, and put back to its original position, there will be color differences. As shown in Fig. \ref{fig:inpaint}, there are some tiny holes in the combination junction, which makes the overall smoothness of the result poor. To this issue, we use \cite{ulyanov2018deep} to repair these tiny holes so that the result has good smoothness.

\begin{figure}[tb]
\begin{center}
    \includegraphics[width=7cm]{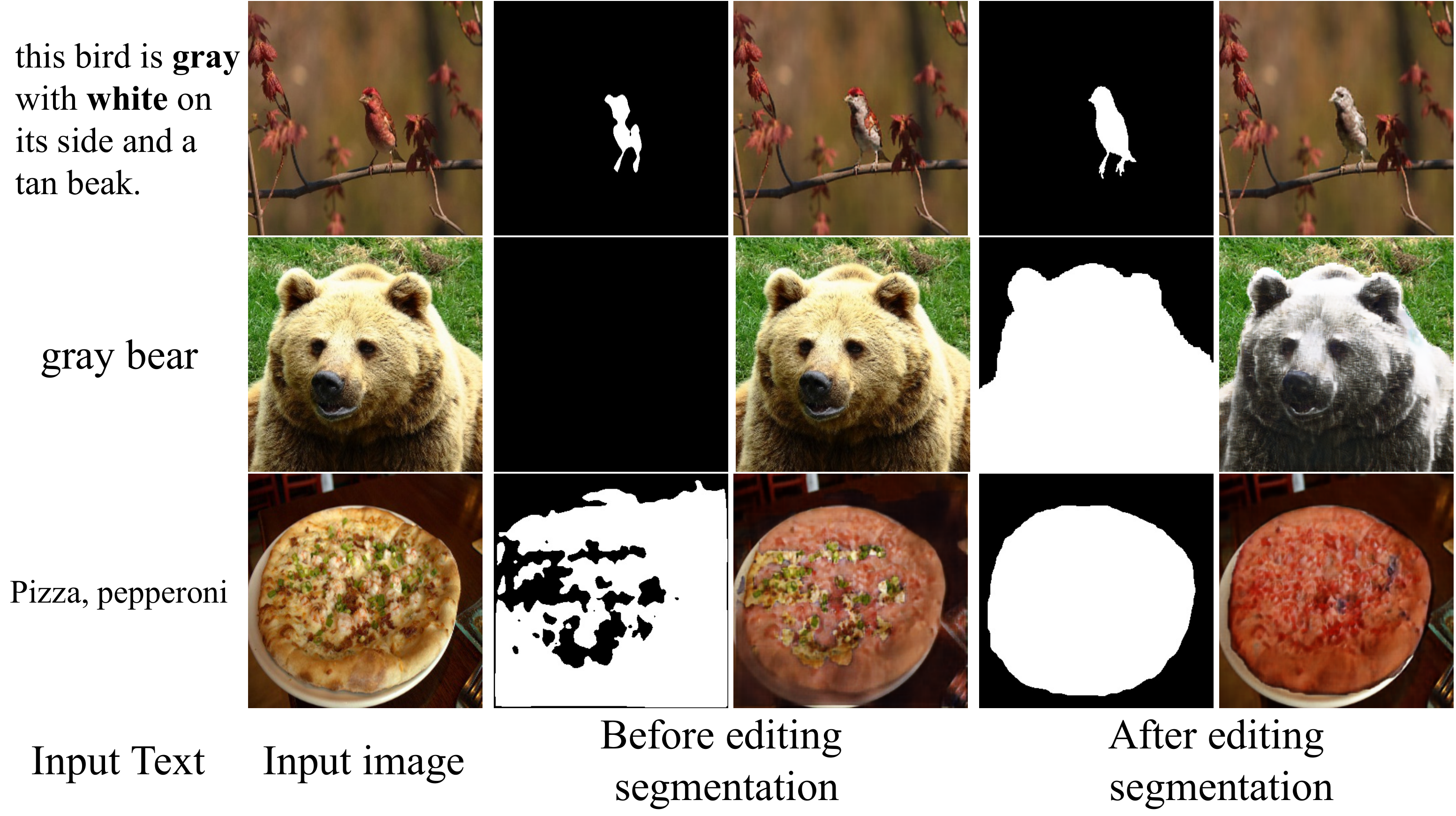}
\end{center}
    \caption{The qualitative result of before and after modifying the segmentation map.}
\label{fig:mis_seg1}
\end{figure}
\begin{figure}[tb]
\begin{center}
    \includegraphics[width=7cm]{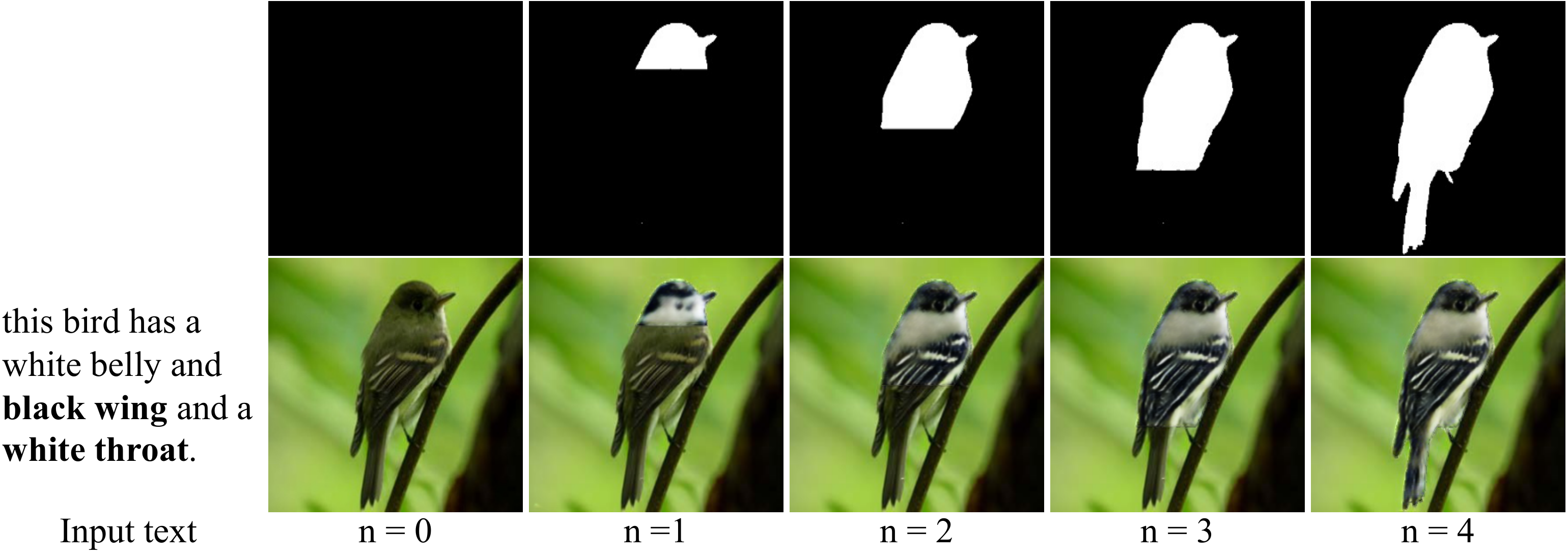}
\end{center}
    \caption{The results of modifying the $n$ times segmentation using the interface.}
\label{fig:n_times_modify}
\end{figure}

\textbf{Improvement from editing segmentation.} Without the user interface to edit segmentation, there are two cases where the text-irrelevant content has penetrated text-relevant content and vice versa.
As shown in Fig. \ref{fig:mis_seg1}, in both cases mentioned above, the image is edited more correctly after the correction than before.
This indicates that the user interface makes it possible to improve the accuracy of image manipulation.
In addition, Fig. \ref{fig:overview} shows an example the image has more than one bird in the image, but the text only describes one bird. In existing works, two birds are forced to change when the edit to match the text description.
However, by using a user interface that interactively modifies the segmentation, we can focus on the specific bird to edit.
Fig. \ref{fig:n_times_modify} shows the improvement of accuracy to modify the segmentation map recursively. In our model, it is possible to modify the segmentation map again even after making the segmentation fail. Therefore, if the generated image does not match the user's desires, we can generate a more accurate image by redoing it many times.
\section{Conclusion}
\label{sec:page}
In this paper, we propose a novel image manipulation method that interactively edits an image using complex text instructions.
By introducing semantic segmentation, it segments the text-relevant and text-irrelevant content of the input image so that the model can only modify the text-relevant content and maintain the text-irrelevant content.
Furthermore, we introduce a super-resolution network and an inpainting network to achieve more operations (such as object enlarging and dwindling, background replacement, etc.) and generate more realistic manipulation results.
Moreover, we propose a user interface that can edit the segmentation map interactively so that the user can obtain satisfactory manipulation results.
Experimental results show that our method outperforms the state-of-the-art models both quantitatively and qualitatively, and also demonstrate that our work has good application value. 

\section{Acknowledgement}
\label{sec:acknowledgement}
This Research is supported by the Joint Research Project of Young Researchers of Hosei University in 2021.

{\small
\bibliographystyle{ieee_fullname}
\bibliography{egbib}
}

\end{document}